\documentclass[10pt,twocolumn,letterpaper]{article}

\usepackage[pagenumbers]{cvpr} 
\makeatletter
\@namedef{ver@everyshi.sty}{}
\makeatother
\usepackage{graphicx}
\usepackage{amsmath}
\usepackage{amssymb}
\usepackage{booktabs}

\usepackage{graphicx}
\usepackage{xcolor}
\usepackage{colortbl}
\usepackage{nicematrix}

\usepackage{textcomp}
\usepackage{stfloats}
\usepackage{url}
\usepackage{verbatim}
\usepackage{graphicx}
\usepackage{cite}
\usepackage{tikz}
\usepackage{comment}
\usepackage{amsmath,amssymb} 
\usepackage{color}
\usepackage{tabularx,lipsum}

\usepackage{wasysym}
\usepackage{graphics} 
\usepackage{epsfig} 
\usepackage{mathptmx} 
\usepackage{color} 
\usepackage{tabularx}
\usepackage{makecell}
\usepackage{multirow}
\usepackage{mwe,lipsum}
\usepackage{array,multirow}
\usepackage{float}
\usepackage{eucal}
\usepackage{comment}
\usepackage{pifont}

\usepackage{arydshln}

\newcommand{\cmark}{\ding{51}}
\newcommand{\xmark}{\ding{55}}

\definecolor{ao}{rgb}{0.0, 0.5, 0.0}

\newcommand{\mysquare}[1][black]{\small\textcolor{#1}{\ensuremath\blacksquare}}
\newcommand{\mycirc}[1][black]{\small\textcolor{#1}{\ensuremath\bullet}}
\newcommand{\mytriangle}[1][black]{\small\textcolor{#1}{\ensuremath\blacktriangle}}

\usepackage[pagebackref,breaklinks,colorlinks]{hyperref}

\usepackage[capitalize]{cleveref}
\crefname{section}{Sec.}{Secs.}
\Crefname{section}{Section}{Sections}
\Crefname{table}{Table}{Tables}
\crefname{table}{Tab.}{Tabs.}

\def\cvprPaperID{3230} 
\def\confName{CVPR}
\def\confYear{2023}

\begin{document}

\title{Source-free Depth for Object Pop-out}

\author{Zongwei Wu$^{1,2,3}$\quad Danda Pani Paudel$^{1,4}$ \quad Deng-Ping Fan$^{1}$\thanks{Corresponding Author: Deng-Ping Fan (dengpfan@gmail.com)}
    \quad Jingjing Wang$^{5}$  \quad Shuo Wang$^{1}$ \\ Cédric Demonceaux$^{2,6}$ \quad Radu Timofte$^{3}$ \quad Luc Van Gool$^{1,4}$ \\
    \small $^{1}$ CVL, ETH Zurich \quad 
    $^2$ University of Burgundy, CNRS, ICB  \quad  $^3$ Computer Vision Lab, CAIDAS \& IFI, University of Wurzburg \\ 
    \small  $^4$ INSAIT, Sofia University \quad  \small  $^5$ AUST \quad 
     $^6$ University of Lorraine, CNRS, Inria, Loria \\
}

\maketitle

\begin{abstract}
  Depth cues are known to be useful for visual perception. However, direct measurement of depth is often impracticable. Fortunately, though, modern learning-based methods offer promising depth maps by inference in the wild. In this work, we adapt such depth inference models for object segmentation using the objects' ``pop-out'' prior in 3D. The ``pop-out'' is a simple composition prior that assumes objects reside on the background surface. Such compositional prior allows us to reason about objects in the 3D space. More specifically, we adapt the inferred depth maps such that objects can be localized using only 3D information. Such separation, however, requires knowledge about contact surface which we learn using the weak supervision of the segmentation mask. Our intermediate representation of contact surface, and thereby reasoning about objects purely in 3D, allows us to better transfer the depth knowledge into semantics.  The proposed adaptation method uses only the depth model without needing the source data used for training, making the learning process efficient and practical. Our experiments on eight datasets of two challenging tasks, namely salient object detection and camouflaged object detection, consistently demonstrate the benefit of our method in terms of both performance and generalizability. The source code is publicly available at \url{https://github.com/Zongwei97/PopNet}. 

\end{abstract}

\section{Introduction}




    

\label{sec:intro}
The 3D knowledge of the scene is long known to be complementary to the task of visual perception~\cite{howard2012perceiving, silberman2012NYUV2,fu2019dual,wu2020depth,zhou2021rgbdsurvey}. Often in practice, though, visual perception needs to be carried out using only 2D images. Given multiple images, 3D geometry may be recovered using the structure-from-motion techniques~\cite{kopf2021rcvd,zhang2021consistent,schonberger2016structure,cvd}. Such inversion, however, is not compatible when only a single image is available. Under such circumstances, image inversion to depth map is usually done using learning-based methods~\cite{monodepth,ranftl2019midas,ranftl2021vision,wu2022toward,miangoleh2021boosting}, which have shown unparalleled success in the recent years. Unfortunately, the learning-based methods may not offer high-quality depth maps due to the generalization deficiency across domains \cite{modality2021wu,xiang2021exploring}. 

Despite poor generalization, the knowledge gained in one domain is shown to be useful in other close-by domains. This utility is harnessed by performing the so-called domain adaptation (DA)~\cite{cardace2022plugging,zhao2019geometry,atapour2018real,saha2021learning,tonioni2019unsupervised,dong2021confident}. In fact, it has been shown recently that DA methods can efficiently transfer knowledge using only the prediction models, \ie, without requiring access to the data where the model is trained -- also known as the source-free domain adaptation (SDA)~\cite{kundu2020universal,yang2021generalized,kim2021domain,liu2021source,xia2021adaptive}. The SDA methods are of gripping interest due to their efficiency and privacy promises.   

\begin{figure}[t]
\centering
\includegraphics[width=\linewidth,keepaspectratio]{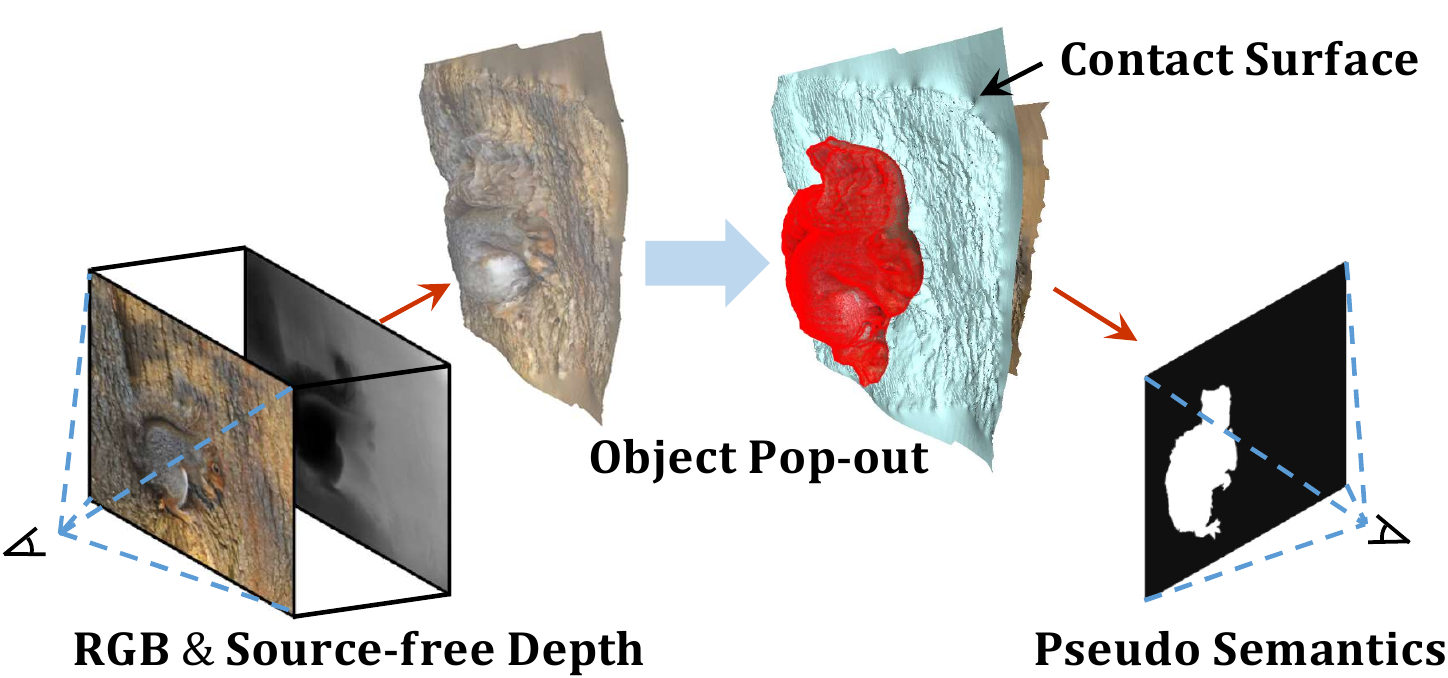}
\vspace{-4mm}
\caption{ Depth to semantics conversion using the \textbf{object pop-out} prior. For input RGB and source-free depth pair, we learn contact surface. The obtained contact surface is then used to separate objects and backgrounds to derive pseudo semantics for supervision.}
\label{fig:1}
\vspace{-2mm}
\end{figure}

\begin{figure*}[t]
\centering
\includegraphics[width=\linewidth,keepaspectratio]{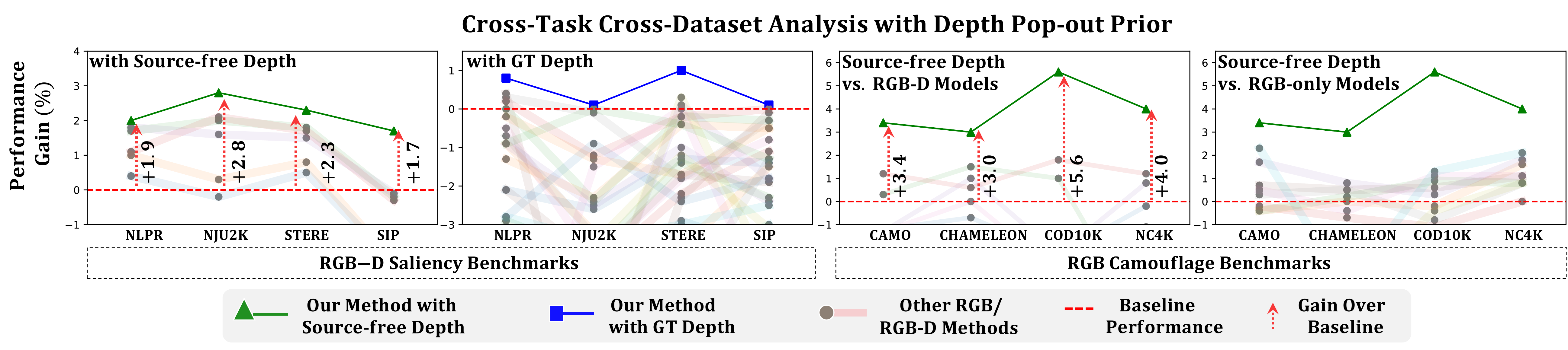}
\vspace{-5mm}
\caption{\textbf{The performance gained} in F-measure using our method over the established baselines. Our method (\mytriangle[ao],\mysquare[blue]) significantly improves baselines on \textbf{8 datasets} and \textbf{2 tasks} (each task on 4 datasets-- SOD: left two; COD: right two). We also compare \textbf{24~methods} (\mycirc[gray]), where our method offers state-of-the-art results despite their task specialization. Note that all methods are connected using lines to illustrate their performance fluctuations across datasets.  Please, refer  Tables~\ref{tab:sod} \&~\ref{tab:cod}  and Section~\ref{abla} for more details and discussions.}
\label{fig:2}
\vspace{-2mm}
\end{figure*}

The most existing SDA methods make one or both of these implicit assumptions: (a)  a similar (as that of the source) supervising task at the target~\cite{kundu2020universal,yang2021generalized,yang2021exploiting}; 
(b) task of discrete (and known) label space~\cite{li2021imbalanced,xia2021adaptive,kurmi2021domain,kundu2022balancing}. The former assumption not only makes the source and target domains easier to compare but also potentially keeps the two domains closer. The latter assumption allows performing SDA by self-training where the discrete labels facilitate reasoning about the model's confidence. The self-training is then performed by boosting the confidence at the target for some picked reliable examples.     

In this work, we aim for source-free transfer of depth knowledge for object detection. Such transfer is desired to assist in locating the object by depth cues and to exploit the depth knowledge despite the domain gap. The addressed problem setting differs from the standard SDA in terms of (a) the source and target tasks' difference; (b) and the continuous label space of the depth. These differences (with standard SDA) make our task at hand very challenging, which is addressed for the first time in this paper, up to our knowledge. To address such a challenging problem, we rely on the ``pop-out'' prior, which allows us to reason about the object's location directly in the 3D space. The ``pop-out'' is a simple composition prior that assumes objects reside \emph{on} the background surface. A graphical illustration of the used ``pop-out'' prior,  using the results obtained by our method, is given in Figure~\ref{fig:1}.

The pop-out prior for image composition was successfully used by \emph{Kang}~\etal in~\cite{kang2009image}. An early work of \emph{Treisman} has provided an in-depth study of such prior in~\cite{treisman1985preattentive}. In this work, we rely on the same compositional foundation of these works and exploit the pop-out before transferring the depth of knowledge across domains. 
Although our motivation comes from these early works, our experimental setup largely differs from theirs. We differ in terms of not only depth knowledge transfer across domains (without source data) but also in target supervision using only semantics. 

\begin{figure*}[t]
\centering
\includegraphics[width=.9\linewidth,keepaspectratio]{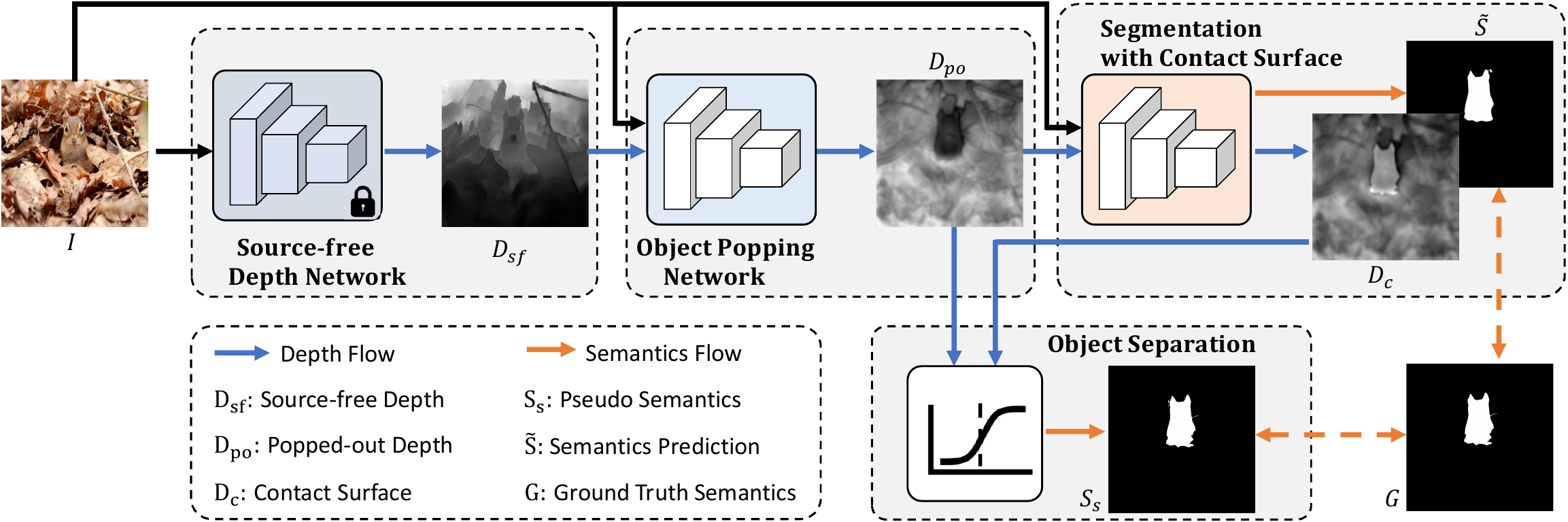}
\put(-65,30){{\color{black}{\footnotesize{$\mathcal{L}_{sep}$}}}}
\put(-39,52){{\color{black}{\footnotesize{$\mathcal{L}_{sem}$}}}}
\caption{Our \textbf{proposed framework}, termed PopNet, is composed of a source-free network, an object popping network, a segmentation with contact surface, and an object separation module. The source-free depth network generates pseudo-depth in an off-the-shelf manner (Section \ref{sfd}). The object popping network converts the source-free depth into the popped-out depth of objects, bridging the cross-domain and cross-task gaps  (Section \ref{opn}). The segmentation network uses this depth to estimate the object's mask and contact surface (Section \ref{scs}). The object separation module then converts popped-out depth to the second mask of objects using the contact surface (Section \ref{os}). We compare both semantic masks against ground truth to supervise the whole pipeline in an end-to-end manner.}
\label{fig:diagram}
\vspace{-2mm}
\end{figure*}

The proposed method exploits the source-free depth to map it into a space where objects in depth stand out better against the background. This mapping is used for object and background separation using a learned contact surface between them. Such separation allows us to derive the semantic masks which can be directly compared against the ground truth for supervision. Using this supervision at the target, we can minimize the domain and task gap between the source and target. The overall framework of our method that performs cross-task cross-domain knowledge transfer by using the intermediate representation of the pop-out space is shown in Figure~\ref{fig:diagram}. As can be seen, we first leverage an object popping network to encourage the object to jump out from the source-free depth. Then, we introduce another network \ie, the segmentation with contact surface, to localize the object and predict the contact surface. These learning modules are jointly trained in an end-to-end manner, transferring the source-free depths into intermediate representations which are adapted to the target task, \ie, object detection. To evaluate the proposed method, we conducted exhaustive experiments on eight datasets of two challenging tasks, namely salient object detection and camouflaged object detection. In both tasks, our method significantly improves the established baselines and offers state-of-the-art results at the same time, whose overview can be seen in Figure~\ref{fig:2}.
The major contributions are as follows:

\begin{itemize}
\setlength{\itemsep}{0pt}
\setlength{\parsep}{0pt}
\setlength{\parskip}{0pt}
\vspace{-2mm}
\item Our problem of transferring source-free depth knowledge across domains and tasks is practical and novel.
    \item Our method relies on our 
    object pop-out prior for visual understanding, which is simple and effective.
    \item Results of our method in two different tasks are significantly better than the baselines and existing models.
\end{itemize}

\section{Related Works}
\label{sec:related}

\noindent \textbf{Source-free Adaptation:} Knowledge transfer by domain adaptation without access to the source data has recently gathered vast interest~\cite{kundu2020universal,yang2021generalized,kim2021domain,liu2021source,xia2021adaptive, agarwal2022unsupervised}, due to the privacy, practicality, and efficiency reasons. We observe accessing the source data for transferring the depth knowledge learned by the off-the-shelf knowledge models~\cite{monodepth,ranftl2019midas} is particularly impractical due to their multi-stage training on various datasets. The existing source-free adaptation methods either use  generative~\cite{li2020model,kurmi2021domain,kundu2020universal}, pseudo-label~\cite{kim2021domain,wang2022continual,liang2020we}, or other customized~\cite{sahoo2020unsupervised,yang2020unsupervised} approaches.
In this work, we use the pseudo-label-based approach. However, using existing methods is not straightforward in our setting due to the tasks' difference between source and target. With our pop-out technique that provides the pseudo semantics, the source-free depth is better transferred across tasks.


\noindent \textbf{Salient Object Detection (SOD):} Saliency detection aims to detect and segment the most prominent region within an image that visually attracts human attention~\cite{fan2022salient}. A number of works have shown that saliency can be an auxiliary step for different vision tasks such as object tracking~\cite{zhou2021saliency}, object detection~\cite{song2021exploiting}, \etc. Conventional saliency works are unimodal, \ie, they only require RGB images as input. In generic and common settings, RGB-based models~\cite{liu2019simple,zhang2017learningsingle,wu2019cascaded} have already achieved very promising results. More recently, several works~\cite{zhao2020depth,piao2020a2dele,ji2020accurate,fan2019rethinkingd3,wu2021mobilesal,modality2021wu,zhouiccvspnet} exploit depth maps as additional clues to the 3D geometry since the depth can provide more truthful information on the object boundary as well as the scale awareness. These 3D features further improve the detection accuracy and performance in challenging scenarios~\cite{Li_2021_HAINet,CDINet,jin2021cdnet,wu2022robust}.


\noindent \textbf{Camouflaged Object Detection (COD):} Camouflage detection aims to find the preying object within an image. For computer vision society, primary works~\cite{li2021uncertainty,pang2022zoom,fan2020camouflaged} often compare COD with SOD. A number of works~\cite{fan2020camouflaged,fan2021concealed,yang2021uncertainty,zhai2022deep} have shown that simply extending saliency models~\cite{wu2019cascaded} on COD will lead to undesired results, which is mainly caused by the nature of target object, \ie, concealed or prominent. Hence, to constrain the attention on the concealed objects, several works come up with different perceptual systems that mimic human behavior vis-a-vis camouflaged objects, such as three-stage localize-segment-rank strategy~\cite{lv2021simultaneously}; iterative refinement~\cite{jia2022segment}, which is similar to repeatedly looking on the images; zooming into possible regions~\cite{pang2022zoom,sun2021c2fnet}. Others~\cite{zhu2021inferring,ji2022gradient,sun2022boundary,zhu2022can,zhong2022detecting,zhang2022preynet,lu2022depth,zhang2021depth} deeply explore the texture difference with the help of the gradient~\cite{ji2022gradient}, frequency~\cite{zhong2022detecting}, edges~\cite{sun2022boundary,zhu2022can}, and probability~\cite{li2021uncertainty, yang2021uncertainty}.

The latest psychological studies~\cite{cuthill2005disruptive,el2021visual} have shown that human perception can naturally benefit from the depth cues to understand the scene: (A) the smooth variation within the object can contribute to alleviating the fake edges and preserving the object structure; (B) the depth discontinuity on the object boundary can make the segmentation easier. Inspired by these observations, we aim to explore the source-free depth for both SOD and COD tasks. To tackle the domain gap for source-free depth maps, we propose to jointly finetune the source-free depth together with the semantic network in an end-to-end manner, with both self-supervised loss and weak semantic supervision.



\section{Proposed PopNet}
\label{sec:methods}
Given an input RGB image $I$ with size $I \in \mathbb{R}^{3\times H \times W}$, where $H$ and $W$ are the spatial resolutions, \ie, height and width, of the image, our objective is to predict the semantic mask $\Tilde{S} \in \mathbb{R}^{H \times W}$ for object detection. As shown in Figure~\ref{fig:diagram}, the input image $I$ is firstly fed into a frozen-weight depth network to generate the source-free depth $D_{sf}\in \mathbb{R}^{H \times W}$ (Section \ref{sfd}). Then the mimicked multi-modal images are fed together into the depth popping network to compute the intermediate popped-out depth $D_{po}\in \mathbb{R}^{H \times W}$ (Section~\ref{opn}). This intermediate representation, as well as the RGB image $I$, is later processed by the segmentation network and transformed into a contact surface $D_c\in \mathbb{R}^{H \times W}$ and a semantics prediction  $\Tilde{S}\in \mathbb{R}^{H \times W}$ (Section~\ref{scs}). On the one hand, the semantics prediction is directly supervised by the ground truth mask $GT$, denoted as $G$, which is similar to conventional segmentation supervision. On the other hand, we further explore the contact surface to pop the object out of the background by means of our object separation module (Section~\ref{os}). This transfers the geometric cues into pseudo semantics and leads to another level of supervision. 


\subsection{Source-free Depth Network}
\label{sfd}

In a practical setting, the GT depth is not always available. Therefore, we generate the source-free depth $D_{sf}$ in an off-the-shelf manner to mimic the multi-modal input. We choose the state-of-the-art DPT model~\cite{ranftl2021vision} with frozen weights as our source-free depth network, which offers us promising depth at the target. This choice is made upon its generalization capability as suggested in~\cite{ranftl2019midas}. To obtain the depth map with the highest quality possible, by enhancing local details, we apply the boosting method~\cite{miangoleh2021boosting} together with DPT. Despite the plausible results achieved by learning-based methods, the obtained source-free depth does not always offer high-quality geometric cues due to the domain gap. Therefore, we leverage the geometric and semantic priors to jointly finetune the source-free depth.

\subsection{Object Popping Network}
\label{opn}

\begin{figure}[t]
\centering
\includegraphics[width=\linewidth,keepaspectratio]{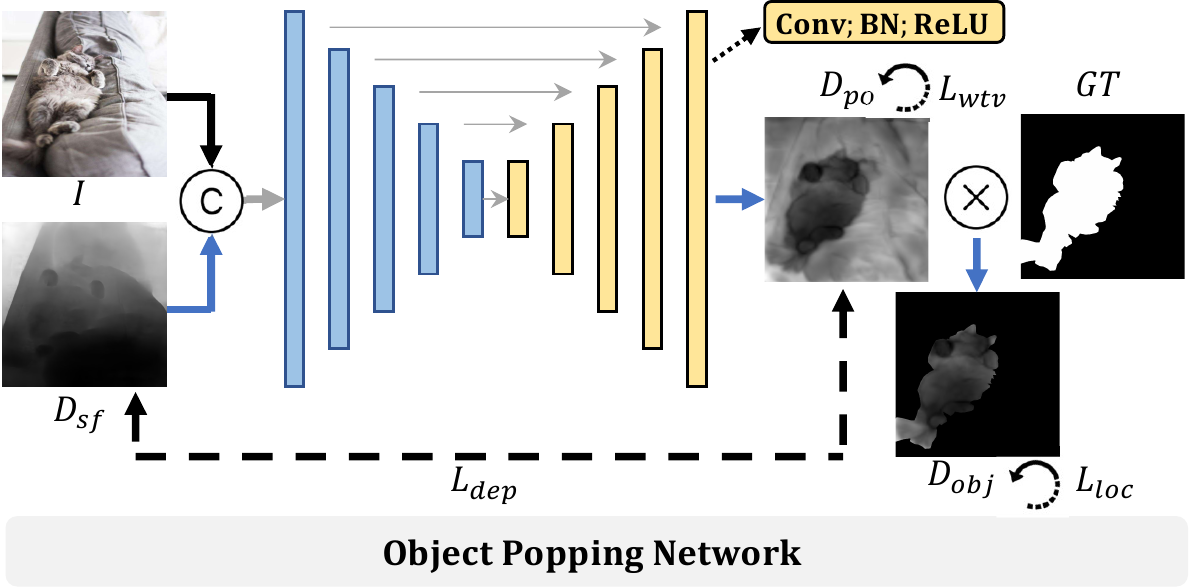}
\vspace{-5mm}
\caption{Our \textbf{object popping network} maps RGB-D inputs into popped-out depths. This network is supervised using a combination of structure preserving, local depth smoothing, and depth edge sharpening losses $\mathcal{L}_{dep}$, $\mathcal{L}_{loc}$, and $\mathcal{L}_{wtv}$, respectively.}
\label{fig:depthpop}
\vspace{-2mm}
\end{figure}

\noindent \textbf{Network Architecture:}
We build a depth popping network to refine/smooth the source-free depth. The popping network follows the encoder-decoder design with skip connection as shown in Figure~\ref{fig:depthpop}. In our case, we simply concatenate RGB and source-free depth at the input side to form a 4-channel input and feed them into the popping network. The encoder extracts semantic cues and generates five-scale outputs. Our decoder is composed of Conv2D, BN, ReLU, and upsampling layers. Following U-Net~\cite{ronneberger2015unet}, we build a skip connection through simple addition. 

\vspace{0.5mm}
\noindent \textbf{Structure Preserving:} To supervise our popping network, we first guide the depth refinement with the help of generated pseudo-depth. We only leverage structural similarity since we aim to detect, preserve, and extract object structure from the intermediate representation. We use the following SSIM loss \cite{monodepth} for structural similarity.
\begin{equation}\label{eq:smim}
    \mathcal{L}_{dep} = SSIM(D_{po}, D_{sf}).
\end{equation}

\noindent \textbf{Local Depth Smoothing:}  In addition to the pseudo-depth supervision, we propose two losses to constrain the depth also by semantics. We assume that the objects' structure should be distinguishable from the background, \ie, it should be smooth within the object region and sharp on the bounding pixels. Hence, we propose to leverage the weak semantic cues together with the geometric priors. Specifically, we first introduce a local smoothness loss. The locality is defined by the ground truth semantics $G$. Technically, we mask out background pixels with element-wise multiplication to suppress the inactive area through $D_{obj} = D_{po} \otimes G$. Let $\nabla_x$ and $\nabla_y$ be the Sobel operations. Then, our local loss $\mathcal{L}_{loc}$ is expressed as:
\begin{equation}
\begin{split}
    \overrightarrow{n(p)} =(-\nabla_x (D_{obj}(p));\  -\nabla_y (D_{obj}(p));\  1 ); \\
    \mathcal{L}_{loc} = \sum_{p}\sum_{q\in \mathcal{N}(p)} 1- cosine(\overrightarrow{n(p)};\   \overrightarrow{n(q)}),
\end{split}
\end{equation}
where $\overrightarrow{n}$ stands for the normal, $p$ denotes the pixels within the object region, $\mathcal{N}(p)$ is the neighboring pixels, and $cosine$ is the cosine similarity between two vectors. This way, our local loss only works on the object area, making the object structure consistent within the target region. Applying local smoothness loss reduces the depth noise at the object level.

\vspace{0.5mm}

\noindent \textbf{Depth Edge Sharpening:} In addition to the local smoothness, we also use edge sharpening. 
The edge sharpening loss is formulated as a weighted total variation.   
For this, we first compute the edge-aware weight $w(p)$ at any pixel $p$ as,
\begin{equation}
w(p) = 
\begin{cases} 
   w_0 ,& \text{if $ \nabla_x (G(p))^2 + \nabla_y (G(p))^2 \neq 0$}, \\
  w_0 + \gamma, & \text{otherwise},
\end{cases}
\end{equation}
where $w_0$ is a pre-defined non-zero weight and $\gamma$ is an additional weight for boundary pixels. In our setting, we choose $w_0$ as the normalized (by the image size) count of the boundary pixels, and we set $\gamma = 0.5$. 
We adopt the square form such that the large gradients play more important roles. Our weighted total-variation loss is given by,
\begin{equation}
    \mathcal{L}_{wtv} = \sum_{p}\sum_{q\in \mathcal{N}(p)}  w(p) \cdot ||D_{po}(p) - D_{po}(q)||_2.
\end{equation}
Our weighted total-variance loss differs from the conventional edge loss, due to our weighting function. More specifically, our weighting function relies on semantic boundaries, unlike the commonly used image gradients \cite{nerfren,hu2019revisiting}. Our motivation for using semantic boundaries instead of image gradients comes from our interest in performing object detection under challenging conditions, such as camouflaged objects. In such cases, image gradients may result in misleading weights. At first glance, our loss function seems to be similar to semantic-guided depth estimation methods \cite{xian2020structure,li2023learning,chen2019towards}. However, \cite{xian2020structure} uses GT depths, and \cite{li2023learning,chen2019towards} use multi-frames for supervision.  Unfortunately, such supervision is not possible in our setup. Note that we exploit single-view source-free depth while only using semantic ground truth for supervision. Furthermore, we transfer the source-free depths' knowledge despite their domain gap. Existing regularizations between depth and semantics are already exploited in our method, which is shown to be complimentary to our pop-out prior. We argue that our network architecture that exploits the pop-out prior is not trivial while simultaneously being generic and easy to use.

The  total objective function $\mathcal{L}_{pop}$ to supervise our object popping network is given by,
\begin{equation}
\mathcal{L}_{pop} = \mathcal{L}_{dep} + \lambda_1  \cdot  \mathcal{L}_{loc} + \lambda_2 \cdot  \mathcal{L}_{wtv},
\end{equation}
where $\lambda_1$ and $\lambda_2$ are the hyperparameters.

\subsection{Segmentation with Contact Surface}
\label{scs}

\begin{figure}[t]
\centering
\includegraphics[width=\linewidth,keepaspectratio]{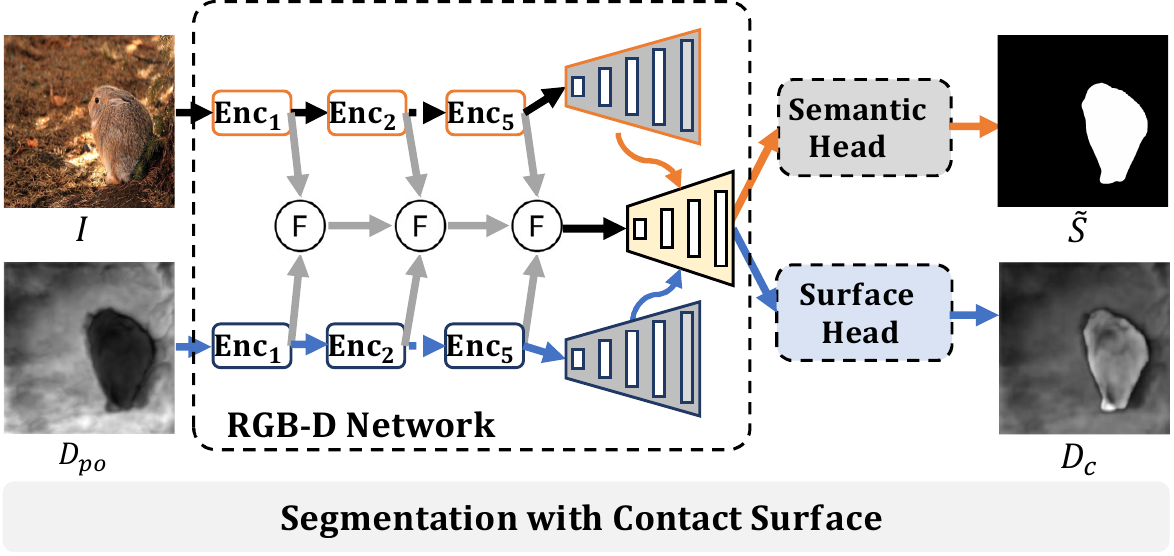}
\vspace{-5mm}
\caption{Our \textbf{segmentation network} uses a basic RGB-D three-stream network~\cite{zhouiccvspnet}. In addition to the conventional semantic head, we learn to predict pixel-wise \textbf{contact surface}. The contact surface is later used to transfer depth knowledge to semantics.}
\label{fig:segmentation}
\vspace{-2mm}
\end{figure}

The smoothness and edge losses encourage homogenizing the object structure, making it noticeable from the background.  We now aim to further enlarge the object-background distance to make the object structure jump out.
Specifically, we use an RGB-D segmentation network as shown in Figure~\ref{fig:segmentation}. The main component of our segmentation network is a three-stream RGB-D network with some fusion design. In our setting, we choose~\cite{zhouiccvspnet} as our baseline since it is one of the SOTA RGB-D methods for saliency detection. We add a surface head to learn the depth of the contact surface $D_c$, which has the same resolution as the input depth $D_{po}$.  Our surface head is composed of ConvLayer (Conv2D, BN, ReLU) and a Conv2D, which first decodes the feature maps and then transfers them into a 1-D map.

\subsection{Object Separation}
\label{os}
Using the previously discussed contact surface, in this section, we aim to separate the object from its background. At this point, we make an assumption that pixels in front of the contact surface belong to objects. The remaining pixels belong to the background. This assumption allows us to explicitly transfer the 3D knowledge into 2D semantics.
Let the predicted depth of the contact surface be $D_c \in \mathbb{R}^{H \times W}$.  We obtain the the pseudo semantics $ S_s$, using the popped-out depth $D_{po}$ and surface's depth $D_c$, as:
\begin{equation}
  S_s = sigmoid(\sigma  \cdot  (D_{po} - D_c)), \\
\end{equation}
where $\sigma$ is a scalar value that controls the slope of the sigmoid function. In our experiments, we use $\sigma=10$ to perform soft-thresholding, mimicking the desired hard one for binary outputs. Such soft thresholding facilitates the gradient back-propagation required for training. 
Finally, we minimize the gap between the pseudo semantics $ S_s$ and the GT semantics $G$ with binary cross-entropy (BCE): 
\begin{equation}
\mathcal{L}_{sep} = BCE(S_s, G).
\end{equation}

\subsection{Overall Loss Function}
Both of our trainable modules, \ie, \textit{object popping} and \textit{segmentation networks}, 
of our framework (Figure~\ref{fig:diagram}) are trained in an end-to-end manner. 
Therefore, the overall loss function consists of three parts: our depth popping loss $\mathcal{L}_{pop}$, our object separation loss $\mathcal{L}_{sep}$, and the conventional semantic loss $\mathcal{L}_{sem}$ from the RGB-D baseline network. The total loss $\mathcal{L}_{total}$ used for training is given by,
\begin{equation}
\mathcal{L}_{total} = \mathcal{L}_{pop} + \alpha_1  \cdot  \mathcal{L}_{sep} + \alpha_2  \cdot  \mathcal{L}_{sem},
\end{equation}
where $\alpha_1$ and $\alpha_2$ are hyperparameters. 

\noindent \textbf{Remarks:} Our losses work in a complementary manner. The $\mathcal{L}_{pop}$ plays the same role as a smooth filter as in image smoothing. It removes the noisy depth response due to the domain gap while preserving the object structure with the help of a weak semantic label. Hence, the smoothed background becomes less informative while the object region becomes uniform, making it easily detectable. These functionalities contribute to transferring the source-free depth into popped-out depth, as desired. Such a process brings objects above the background surface, despite their distance from the camera and other distracting surfaces. 
The $\mathcal{L}_{sep}$, on the other hand, fully benefits from the ``pop-out'' prior to segment the object from the background. With the help of the learned contact surface, this loss enlarges the foreground-background distance by pulling them in opposite directions. Such pulling results in a binary-like mask, which effectively bridges the gap between geometric knowledge and semantics.  Finally, both depth-transferred and learned semantics are compared to the ground truths for supervised training.

\begin{table*}[t]
\footnotesize
\setlength\tabcolsep{4.5pt}
\renewcommand{\arraystretch}{1}
\begin{center}
\caption{Quantitative comparison on RGB-D SOD datasets.  $\uparrow$ ($\downarrow$) denotes that the higher (lower) is better.  We use the Mean Absolute Error ($M$), max F-measure ($F_m$), S-measure ($S_m$), and max E-measure ($E_m$) as evaluation metrics. G.D. stands for GT Depth.  \textbf{Bold} denotes the best performance.}
\label{tab:sod}
\begin{tabular*}{\linewidth}{ll||llll|llll|llll|llll}

\hline

\hline

\multirow{2}{*}{G.D. \quad Public.}  & Dataset &\multicolumn{4}{c}{NLPR~\cite{peng2014rgbd}} &  \multicolumn{4}{c}{NJUK~\cite{ju2014depth}} & \multicolumn{4}{c}{STERE~\cite{niu2012leveraging}} & \multicolumn{4}{c}{SIP~\cite{fan2019rethinkingd3}}\\
\cline{3-6} \cline{7-10} \cline{11-14} \cline{15-18} 

& Metric & 
        $M\downarrow$ & $F_{m}\uparrow$ &  $S_m\uparrow $ & $E_m\uparrow$ &
        $M\downarrow$ & $F_{m}\uparrow$ &  $S_m\uparrow $ & $E_m\uparrow$ &
        $M\downarrow$ & $F_{m}\uparrow$ &  $S_m\uparrow $ & $E_m\uparrow$ &
        $M\downarrow$ & $F_{m}\uparrow$ &  $S_m\uparrow $ & $E_m\uparrow$ \\
\hline

\multicolumn{15}{l}{\textbf{Performance of RGB-D Models Trained with Source-free Depth}} \\

\xmark \quad $MM_{21} $~\cite{Zhang2021DFMNet} & DFM-Net 
                                      &  .027&   .909&   .914&   .944
                                      &  .046&   .903&   .895&   .927
                                      &   .042&   .906&   .903&   .934
                                      &  .067&   .873&   .850&   .891 \\

\xmark \quad $TIP_{22}$ ~\cite{wang2022learning} & DCMF 
                                      &  .027&   .915&   .921&   .943
                                      &  .044&   .908&   .903&   .929
                                      &   .041&   .909&   .907&   .931
                                      &  .067&   .873&   .853&   .893 \\
                                      

\xmark \quad $CVPR_{22}$ ~\cite{jia2022segment} & SegMAR 
                                      &  .024&   .923&   .920&   .952
                                      & .036 & .921 & .909 & .941
                                      &  .037&   .916&   .907&   .936
                                      & .052 & .893 & .872 & .914 \\

\xmark \quad $CVPR_{22}$ ~\cite{pang2022zoom} & ZoomNet 
                                      &  .023&   .916&   .919&   .944
                                      &.037  & .926   & .914  & .940
                                      &  .037&   .918&   .909&   .938
                                      &.054 &.891 &.868 &.909\\
\hdashline
\rowcolor[RGB]{235,235,250}               
\xmark  \quad   \textbf{Ours} & \textbf{PopNet}

                                      &  \textbf{.022}&   \textbf{.925}&   \textbf{.926}&   \textbf{.956} 
                                      &  \textbf{.031}&   \textbf{.931}&   \textbf{.920}&   \textbf{.949}
                                      &  \textbf{.032}&   \textbf{.922}&   \textbf{.916}&   \textbf{.947} 
                                      &  \textbf{.046}&   \textbf{.911}&   \textbf{.885}&   \textbf{.926} \\
                                      

\hline
\multicolumn{15}{l}{\textbf{Performance of RGB-D Models Trained with GT Depth}} \\


                                     
 
%

\cmark \ \; $TIP_{21}$~\cite{zhang2021bilateral} & BIANet
                                       
                                      &  .032&   .888&   .900&   .930 
                                      &  .056&   .878&   .867&   .898
                                      &  .048&   .898&   .895&   .918 
                                      &  .091&   .816&   .802&   .847 \\

\cmark \ \; $TIP_{21}$~\cite{Li_2021_HAINet}& HAINet
                                       
                                      &  .024&   .920&   .924&   .956 
                                      &  .037&   .924&   .911&   .940
                                      &  .040&   .917&   .907&   .938 
                                      &  .052&   .907&   .879&   .917 \\

\cmark \ \;  $TNNLS_{21}$\cite{fan2019rethinkingd3} & D3Net
                                       
                                      &  .029&   .904&   .911&   .942 
                                      &  .046&   .909&   .899&   .927
                                      &  .044&   .902&   .906&   .925 
                                      &  .063&   .880&   .860&   .897 \\ 
                                      
%
                                      
\cmark \ \; $ECCV_{22}$~\cite{lee2022spsn} & SPSN
                                       
                                      &  .023&   .917&   .923&   .956 
                                      &  .032&   .927&   .918&   .949
                                      &  .035&   .909&   .906&   .941 
                                      &  .043&   .910&   .891&   .932 \\   
                                       

                                       

\hdashline
\rowcolor[RGB]{235,235,250}  

\cmark \ \ \    \textbf{Ours} & \textbf{PopNet} 
                                      &  \textbf{.019}&   \textbf{.927}&   \textbf{.932}&   \textbf{.963 }
                                      &  \textbf{.030}&   \textbf{.936}&   \textbf{.924}&   \textbf{.952}
                                      &  \textbf{.033}&   \textbf{.924}&   \textbf{.917}&   \textbf{.947 }
                                      &  \textbf{.040}&   \textbf{.923}&   \textbf{.897}&  \textbf{.937} \\
                                      

\hline

\hline
\end{tabular*}
\end{center}
\vspace{-8mm}

\end{table*}

\section{Results}
\label{sec:results}

\subsection{Experimental Setup}
\noindent \textbf{Dataset Preparation:}  To better illustrate the generalizability of our approach, we evaluate the effectiveness of our approach on both SOD and COD benchmarks. We choose four widely used RGB-D SOD datasets, \ie, NLPR~\cite{peng2014rgbd}, NJUK~\cite{ju2014depth}, STERE~\cite{niu2012leveraging}, and SIP~\cite{fan2019rethinkingd3}, as well as four COD datasets, \ie, CAMO~\cite{le2019camo}, CHAMELEON~\cite{skurowski2018chameleon}, COD10K~\cite{fan2020camouflaged}, and NC4K~\cite{lv2021simultaneously}. For SOD datasets, we conduct experiments with both GT depth and source-free depth. We follow the conventional learning protocol~\cite{ji2021calibrated, zhouiccvspnet, wu2022robust} and use 700 images from NLPR and 1,485 images from NJUK for training. The rest are used for testing. For the unimodal COD dataset, we compare with both RGB COD models and RGB-D SOD models retrained on the COD datasets with the same source-free depth $D_{sf}$.  We follow the conventional training/testing protocol~\cite{pang2022zoom,fan2021concealed,fan2020camouflaged,lv2021simultaneously,jia2022segment} and use 3,040 images from COD10K and 1,000 images from CAMO for training. The rest are used for testing.  

\noindent \textbf{Evaluation Metrics:}  We evaluate the performance with four generally-recognized metrics: Mean Absolute Error ($M$), max F-measure ($F_m$), S-measure ($S_m$), and max E-measure ($E_m$). All the object segmentation masks are trained or downloaded from the official resources. To make a fair comparison, we evaluate the prediction semantics with the standardized evaluation protocol as~\cite{zhouiccvspnet}. 


\noindent \textbf{Implementation Details:}
Our model is implemented based on Pytorch with a V100 GPU. We use the Adam algorithm as an optimizer. The learning rate is initialized to 1$e-$4  and is further divided by 10 every 60 epochs. We set the input resolution to 352$\times$352 resolution for RGB and depth. A detailed comparison with higher resolution can be found in Table \ref{tab:rebzoom}.  During training, conventional data augmentation such as random flipping, rotating, and border clipping are adopted. The training takes around 6 hours for RGB-SOD tasks and 12 hours for COD tasks for 100 epochs.

\begin{table*}[t]
\footnotesize
\setlength\tabcolsep{4.7pt}
\renewcommand{\arraystretch}{1.0}
\begin{center}
\caption{Quantitative comparison on COD datasets. Pseudo stands for source-free depth used for RGB-D methods.}


\label{tab:cod}
\begin{tabular*}{\linewidth}{ll||llll|llll|llll|llll}
\hline

\hline

\multirow{2}{*}{Pseudo \ Public.} & Dataset &\multicolumn{4}{c|}{CAMO~\cite{le2019camo}} &  \multicolumn{4}{c|}{CHAMELEON~\cite{skurowski2018chameleon}} & \multicolumn{4}{c|}{COD10K~\cite{fan2020camouflaged}} & \multicolumn{4}{c}{NC4K~\cite{lv2021simultaneously}}\\
\cline{3-6} \cline{7-10} \cline{11-14} \cline{15-18} 

& Metric & 
        $M\downarrow$ & $F_{m}\uparrow$ &  $S_m\uparrow $ & $E_m\uparrow$ &
        $M\downarrow$ & $F_{m}\uparrow$ &  $S_m\uparrow $ & $E_m\uparrow$ &
        $M\downarrow$ & $F_{m}\uparrow$ &  $S_m\uparrow $ & $E_m\uparrow$ &
        $M\downarrow$ & $F_{m}\uparrow$ &  $S_m\uparrow $ & $E_m\uparrow$ \\
\hline

\multicolumn{15}{l}{\textbf{Performance of RGB COD Models}} \\
       
\xmark  \quad  $CVPR_{20}$ ~\cite{fan2020camouflaged} & SINet
                                       
                                      &  .099&   .762&   .751&   .790
                                      &  .044&   .845&   .868&   .908
                                      &  .051&   .708&   .771&   .832
                                      &  .058&   .804&   .808&   .873 \\

\xmark \quad  $CVPR_{21} $~\cite{lv2021simultaneously}   &  SLSR
                                       
                                      &  .080&   .791&   .787&  .843
                                      &  .030&   .866&   .889&   .938
                                      &  .037&   .756&   .804&   .854
                                      &  .048&   .836&   .839&   .898 \\
\xmark \quad $CVPR_{21}$ ~\cite{zhai2021mutual}   & MGL-R
                                   
                                      &  .088&   .791&   .775&   .820
                                      &  .031&   .868&   .893&   .932
                                      &  .035&   .767&   .813&   .874
                                      &  .053&   .828&   .832&   .876 \\
\xmark \quad $CVPR_{21} $  ~\cite{mei2021camouflaged}  &  PFNet
                                       
                                      &  .085&   .793&   .782&   .845
                                      &  .033&   .859&   .882&   .927
                                      &  .040&   .747&   .800&   .880
                                      &  .053&   .820&   .829&   .891 \\
\xmark \quad $CVPR_{21}$ ~\cite{li2021uncertainty}   & UJSC
                                       
                                      &  \textbf{.072}&   .812&   .800&   .861
                                      &  .030&   .874&   .891&   .948
                                      &  .035&   .761&   .808&   .886
                                      &  .047&   .838&   .841&   .900 \\
\xmark \quad $IJCAI_{21}$ ~\cite{sun2021c2fnet}  & C2FNet
                                       
                                      &  .079&   .802&   .796&   .856
                                      &  .032&   .871&   .888&   .936
                                      &  .036&   .764&   .813&   .894
                                      &  .049&   .831&   .838&   .898 \\
\xmark \quad $ICCV_{21}$~\cite{yang2021uncertainty}   & UGTR 
                                       
                                      &  .086&   .800&   .783&   .829
                                      &  .031&   .862&   .887&   .926
                                      &  .036&   .769&   .816&   .873
                                      &  .052&   .831&   .839&   .884 \\
                                     


\xmark  \quad $CVPR_{22}$~\cite{jia2022segment}   & SegMAR 
                                      &  .080&   .799&   .794&  .857 
                                      &  .032&   .871&   .887&   .935
                                      &  .039&   .750&   .799&   .876
                                      &  .050&   .828&   .836&   .893 \\

\xmark \quad $CVPR_{22}$ \cite{pang2022zoom} & ZoomNet 
                                      &  .074&   .818&   .801&   .858
                                      &  .033&   .829&   .859&   .915
                                      &  .034&   .771&   .808&   .872
                                      &  .045&   .841&   .843&   .893 \\

                                      

                                       
                                      
\hline

\multicolumn{15}{l}{\textbf{Performance of RGB-D Models Retrained with Source-free Depth}} \\

                                       

                                       
                                      
                                       

                                       


\cmark \ \;  $MM_{21}$ ~\cite{CDINet} & CDINet
                                       
                                      &  .100&   .638&   .732&   .766 
                                      &  .036&   .787&   .879&   .903
                                      &  .044&   .610&   .778&   .821 
                                      &  .067&   .697&   .793&   .830 \\

                                       
\cmark \ \; $CVPR_{21}$ ~\cite{ji2021calibrated}  &DCF 
                                      &  .089&   .724&   .749&   .834
                                      &  .037&   .821&   .850&   .923
                                      &  .040&   .685&   .766&   .864
                                      &  .061&   .765&   .791&   .878 \\


\cmark \ \; $ICCV_{21}$ ~\cite{cascaded_cmi} &CMINet
                                       
                                      &  .087&   .798&   .782&   .827 
                                      &  .032&   .881&   .891&   .930
                                      &  .039&   .768&   .811&   .868 
                                      &  .053&   .832&   .839&   .888 \\

\cmark \ \;  $ICCV_{21}$~\cite{zhouiccvspnet}  &SPNet
                                       
                                      &  .083&   .807&   .783&   .831 
                                      &  .033&   .872&   .888&   .930
                                      &  .037&   .776&   .808&   .869 
                                      &  .054&   .828&   .825&   .874 \\
       
\cmark \ \; $TIP_{22}$ ~\cite{wang2022learning} &DCMF                            &  .115&   .737&   .728&   .757 
                                      &  .059&   .807&   .830&   .853
                                      &  .063&   .679&   .748&   .776 
                                      &  .077&   .782&   .794&   .820 \\

                                       
                                      
                                       

\cmark \ \;  $ECCV_{22}$~\cite{lee2022spsn}  &SPSN
                                       
                                      &  .084&   .782&   .773&   .829
                                      &  .032&   .866&   .887&   .932
                                      &  .042&   .727&   .789&   .854 
                                      &  .059&   .803&   .813&   .867 \\
                                       

%
                              
                                       
%

\hdashline

                                       
                                      
\rowcolor[RGB]{235,235,250}
\cmark \ \;   \textbf{Ours} & \textbf{PopNet} 
                                       
                                      &  .073&   \textbf{.821}&   \textbf{.806}&   \textbf{.869} 
                                      &  \textbf{.022}&   \textbf{.893}&   \textbf{.910}&    \textbf{.962} 
                                      &  \textbf{.031}&   \textbf{.789}&   \textbf{.827}&   \textbf{.897} 
                                      &  \textbf{.043}&   \textbf{.852}&   \textbf{.852}&   \textbf{.908} \\  
                                      



\hline

\hline
\end{tabular*}
\end{center}

\vspace{-2mm}

\end{table*}

\subsection{Comparisons}



\begin{figure}[t]
\centering
\includegraphics[width=\linewidth,keepaspectratio]{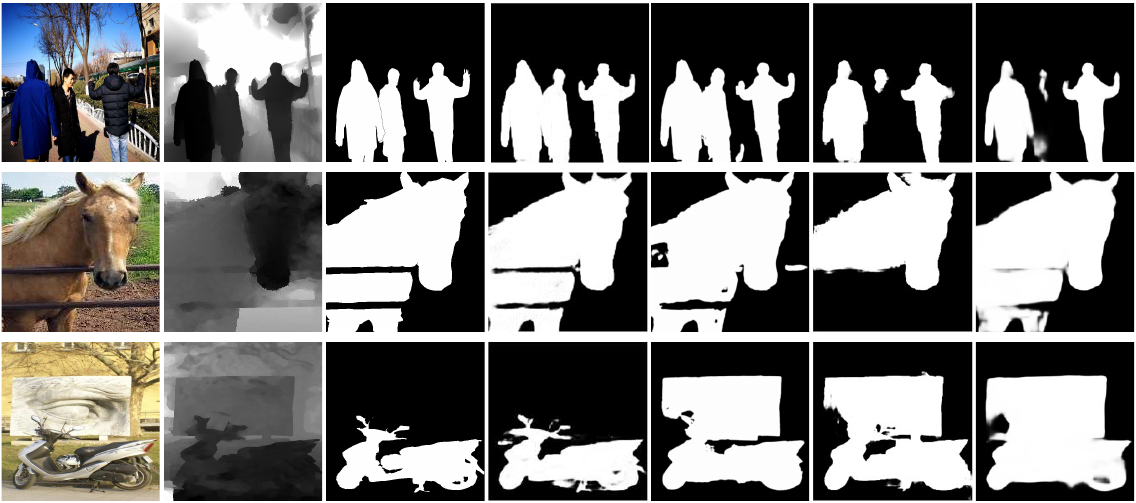}
\vspace{-8mm}
  \put(-110,5){{\color{black}{\small{Image}}}}
  \put(-78,5){{\color{black}{\small{Depth}}}}
  \put(-40,5){{\color{black}{\small{GT}}}}
  \put(-9,5){{\color{black}{\small{Ours}}}}
  \put(23,5){{\color{black}{\small{SPSN}}}}
  \put(54,5){{\color{black}{\small{DSA2F}}}}
  \put(92,5){{\color{black}{\small{DCF}}}}
  \vspace{6mm}
\caption{
\textbf{Qualitative Comparison with GT Depth}. Our method outperforms all counterparts while dealing with multi-objects, large depth variation, and visually-mixed foreground-background.
}
\label{fig:rgbd}
\vspace{-5mm}
\end{figure}
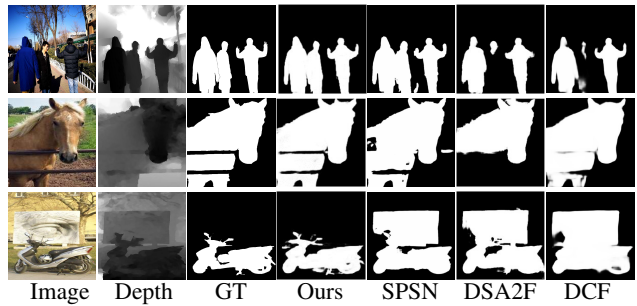

\noindent \textbf{Comparison with RGB-D SOD Models:} We present in Table \ref{tab:sod} the performance on SOD benchmarks with our source-free depth or with GT depth. It can be seen that our model with source-free depth achieves very competitive performance compared to many RGB-D models with GT depths. Our method with GT depth also outperforms the SOTA counterparts. The qualitative comparison can be found in Fig. \ref{fig:rgbd}. Note that we also retrain the SOTA unimodal COD models SegMAR \cite{jia2022segment} and ZoomNet~\cite{pang2022zoom} on the SOD dataset only with RGB images. We show that our model with source-free depth outperforms these counterparts, showing that our method can better generalize across different tasks with favorably better performance.

\noindent \textbf{Comparison with COD Models:} We present in Table \ref{tab:cod} the performance of the most competitive SOTA methods, including task-specific COD models as well as retrained RGB-D SOD models with source-free depth. For a fair comparison, we retrain all RGB-D methods,  SegMAR~\cite{jia2022segment}, and ZoomNet \cite{pang2022zoom} in an end-to-end manner on the same resolution images as ours. Some RGB-only methods perform even better than many RGB-D methods, mainly due to their COD task-specific designs and the lack of ground-truth depths required for RGB-D methods.
It is important to note that competing RGB-only COD methods do not perform favorably on SOD tasks. Please, refer to Table~\ref{tab:sod} \& \ref{tab:rebzoom} to observe their poor cross-tasks generalization.  

\noindent \textbf{Towards Higher Resolution:} Previous studies have shown that the image resolution may influence the model performance \cite{pang2022zoom,xiang2021exploring,mckee2022transfer,zhang2022preynet}. For example, the current SOTA COD method ZoomNet \cite{pang2022zoom} is with main scale $384^2$ and implies the highest resolution of $(384\times1.5)^2=576^2$, as it operates on $0.5\times, 1\times$, and $1.5\times$ scales. To make a fair comparison, we retrain the model with the same resolution ($352^2$ or $512^2$) as ours. We show in Table\ref{tab:rebzoom}-ZoomNet* that the results deteriorate as expected. Compared to these counterparts, our method offers a good trade-off between accuracy and efficiency. More comparisons on SOD and COD benchmarks can be found in the \textcolor{magenta}{supplementary material}.

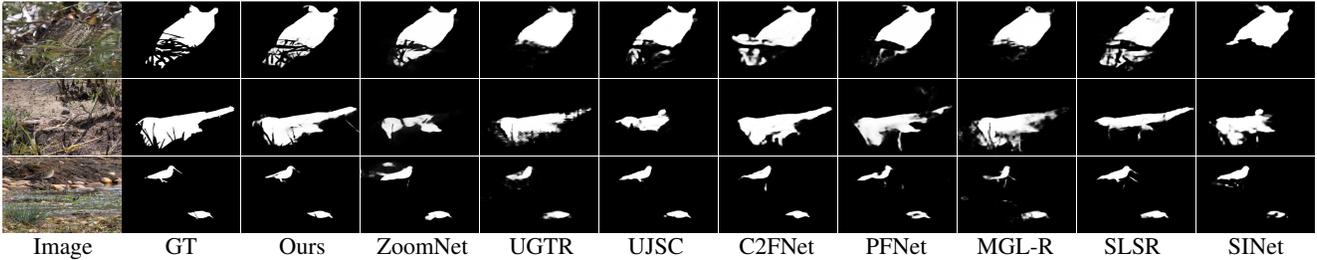
\begin{figure*}[t]
\centering
\includegraphics[width=\linewidth,keepaspectratio]{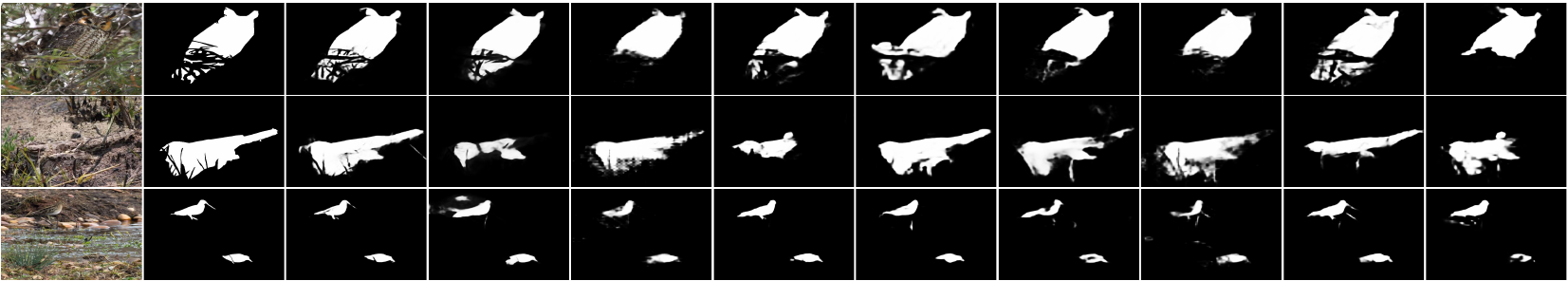}
  \put(-485,-8){{\color{black}{\small{Image}}}}
  \put(-435,-8){{\color{black}{\small{GT}}}}
  \put(-392,-8){{\color{black}{\small{Ours}}}}
 \put(-355,-8){{\color{black}{\small{ZoomNet}}}}
  \put(-305,-8){{\color{black}{\small{UGTR}}}}
  \put(-260,-8){{\color{black}{\small{UJSC}}}}
  \put(-218,-8){{\color{black}{\small{C2FNet}}}}
  \put(-170,-8){{\color{black}{\small{PFNet}}}}
  \put(-128,-8){{\color{black}{\small{MGL-R}}}}
  \put(-80,-8){{\color{black}{\small{SLSR}}}}
  \put(-33,-8){{\color{black}{\small{SINet}}}}
  \vspace{1mm}
\caption{\textbf{Qualitative comparison}. Our method can better preserve the object structure compared to other counterparts, especially while dealing with objects with occlusion (two first rows). Our method performs favorably with multiple objects (last row). Better to zoom in.}
\label{fig:quali}
\vspace{-3mm}
\end{figure*}

\noindent \textbf{Qualitative Comparisons:} Figure~\ref{fig:quali} presents the output of our network on challenging cases. It can be seen while dealing with objects occluded by thin objects ($1^{st} - 4^{th}$ rows), our method can accurately reason about the segmentation masks closer to the GT. We also achieve better performance while dealing with multiple objects (last row). More discussions on multiple objects can be found in Table \ref{tab:rebmultiobj}.

\begin{table}[t]
\scriptsize	
\setlength\tabcolsep{0.35pt}
\renewcommand{\arraystretch}{1.0}
\begin{center}
\caption{End-to-end comparison with different resolutions on SOD benchmarks. Our method with source-free depth generalizes significantly better compared to SOTA COD models.}
\label{tab:rebzoom}
\begin{tabular}{  m{1.6cm} m{0.5cm} m{0.7cm} ||m{.62cm} m{.62cm} m{.62cm}  m{.62cm}|  m{.62cm}  m{.62cm} m{.62cm} m{.62cm} }
\hline

\hline

\hline

\multirow{2}{*}{Model}  & \multirow{2}{*}{Size} &Flops  &\multicolumn{4}{c}{NJUK \cite{ju2014depth}}   & \multicolumn{4}{|c}{SIP \cite{fan2019rethinkingd3}}  \\

\cline{4-11}  & & (G) &$M\downarrow$ & $F_{m}\uparrow$ &  $S_m\uparrow $ & $E_m\uparrow$ & $M\downarrow$ & $F_{m}\uparrow$ &  $S_m\uparrow $ & $E_m\uparrow$\\
\hline
SegMAR \cite{jia2022segment} &$352^2$ & 67.3  & .036 & .921 & .909 & .941  & .052 & .893 & .872 & .914 \\
                                      
ZoomNet \cite{pang2022zoom} &$352^2$ & 167.8 &.037  & .926   & .914  & .940  &.054 &.891 &.868 &.909\\

\hdashline
\rowcolor[RGB]{235,235,250}  

\textbf{Ours}  &$352^2$ & 228.8 &  \textbf{.031}&   \textbf{.931}&   \textbf{.920}&    \textbf{.949} 
                                      &  \textbf{.046}&   \textbf{.911}&   \textbf{.885}&   \textbf{.926}   \\  

\hline

\hline

SegMAR \cite{jia2022segment} &$512^2$ & 142.4 & .035 & .927 & .914 & .943  & .050 & .899 & .878 & .917 \\
                                      
ZoomNet \cite{pang2022zoom} &$512^2$ & 353.4 & .036 &.926 &.915 &.942  &.052 &.895 &.873 &.910 \\

  \hdashline
\rowcolor[RGB]{235,235,250}  
\textbf{Ours}  &$512^2$ & 484.0&  \textbf{.031}&   \textbf{.933}&   \textbf{.922}&    \textbf{.951}  &  \textbf{.044}&   \textbf{.911}&   \textbf{.890}&   \textbf{.927} \\

\hline

\hline

\hline
\end{tabular}
\end{center}
\vspace{-6mm}
\end{table}

\begin{table}[t]
\scriptsize
\setlength\tabcolsep{0.8pt}
\renewcommand{\arraystretch}{1.0}
\begin{center}
\caption{Ablation study on the proposed losses.}
\label{tab:lossabla}
\begin{tabular}{  m{.60cm} m{.60cm} m{.60cm} m{.60cm} ||  m{.61cm}  m{.61cm} m{.61cm} m{.61cm} | m{.61cm}  m{.61cm} m{.61cm} m{.61cm}}
\hline

\hline

\hline

\multirow{2}{*}{$\mathcal{L}_{dep}$}  & \multirow{2}{*}{$\mathcal{L}_{loc}$} & \multirow{2}{*}{$\mathcal{L}_{wtv}$} & \multirow{2}{*}{$\mathcal{L}_{sep}$}  & \multicolumn{4}{c}{SIP~\cite{fan2019rethinkingd3}}  & \multicolumn{4}{|c}{NC4K~\cite{lv2021simultaneously}}  \\

\cline{5-12}

& & &  &  $M\downarrow$ & $F_{m}\uparrow$ &  $S_m\uparrow $ & $E_m\uparrow$ & $M\downarrow$ & $F_{m}\uparrow$ &  $S_m\uparrow $ & $E_m\uparrow$\\

\hline

 - &- &- &-   &  .048&   .903&   .884&   .922
                                      &  .052&   .832&   .832&   .893 \\

 \cmark  &- & -& -  & .046 & .907 & .889 &.925  & .051 & .833 & .839 &.895\\
 
-&  \cmark  & -&-   & .045 & .908 & .893 &.929  & .048 & .837 & .844 &.898\\ 

- &  -&  \cmark & -  & .046 & .906 & .891 &.927  & .050 & .833 & .841 &.894\\ 
  
-&- & - & \cmark  & .043 & .914 & .893 &.933 & .048 & .840 & .848 &.900\\

 \hline
 \cmark & \cmark  & - & - & .044 & .911 & .893 & .928 & .049 & .837 & .844 & .897\\
 \cmark & - &\cmark  & -  & .046 & .909 & .893 & .927 & .046 & .840 & .845 & .898\\
  \cmark & - &  \cmark &  \cmark & \textbf{.040} & .918 & \textbf{.897} & .935 & .045 & .848 & .849 & .904\\
  
 \cmark & \cmark & - & \cmark  &  .042 &.916 &.894 &.931 &  .044 &.850 &.850 &.906\\
 \rowcolor[RGB]{235,235,250}  

  \cmark & \cmark & \cmark & \cmark  &  \textbf{.040}&   \textbf{.923}&   \textbf{.897}&  \textbf{.937}
                                      &  \textbf{.043}&   \textbf{.852}&   \textbf{.852}&   \textbf{.908} \\ 
\hline

\hline

\hline
\end{tabular}
\end{center}
\vspace{-6mm}
\end{table}

\subsection{Ablation Study}
\label{abla}

\noindent \textbf{Loss:} In this section, we conduct experiments on analyzing the effectiveness of the proposed losses. The quantitative results of different loss combinations are provided in Table~\ref{tab:lossabla}. It can be seen that each proposed loss behaves properly, \ie, improving the performance compared to the baseline. More discussions and ablation studies on the hyperparameters can be found in the \textcolor{magenta}{supplementary material}.

\noindent \textbf{Object Popping Network as Plug-in:} Our object popping network can be easily adapted with different encoders and with different existing RGB-D models. For example, with a ResNet-18~\cite{He2016Residual} encoder and convolution-based decoder,  our poping only costs around 12.7M additional learning parameters or 48.7 MB model size. We show in Table \ref{tab:general} that our method can favorably improve performance over the baseline with less than an extra 10\% GFlops. 



\begin{figure}[t]
\centering
\includegraphics[width=\linewidth,keepaspectratio]{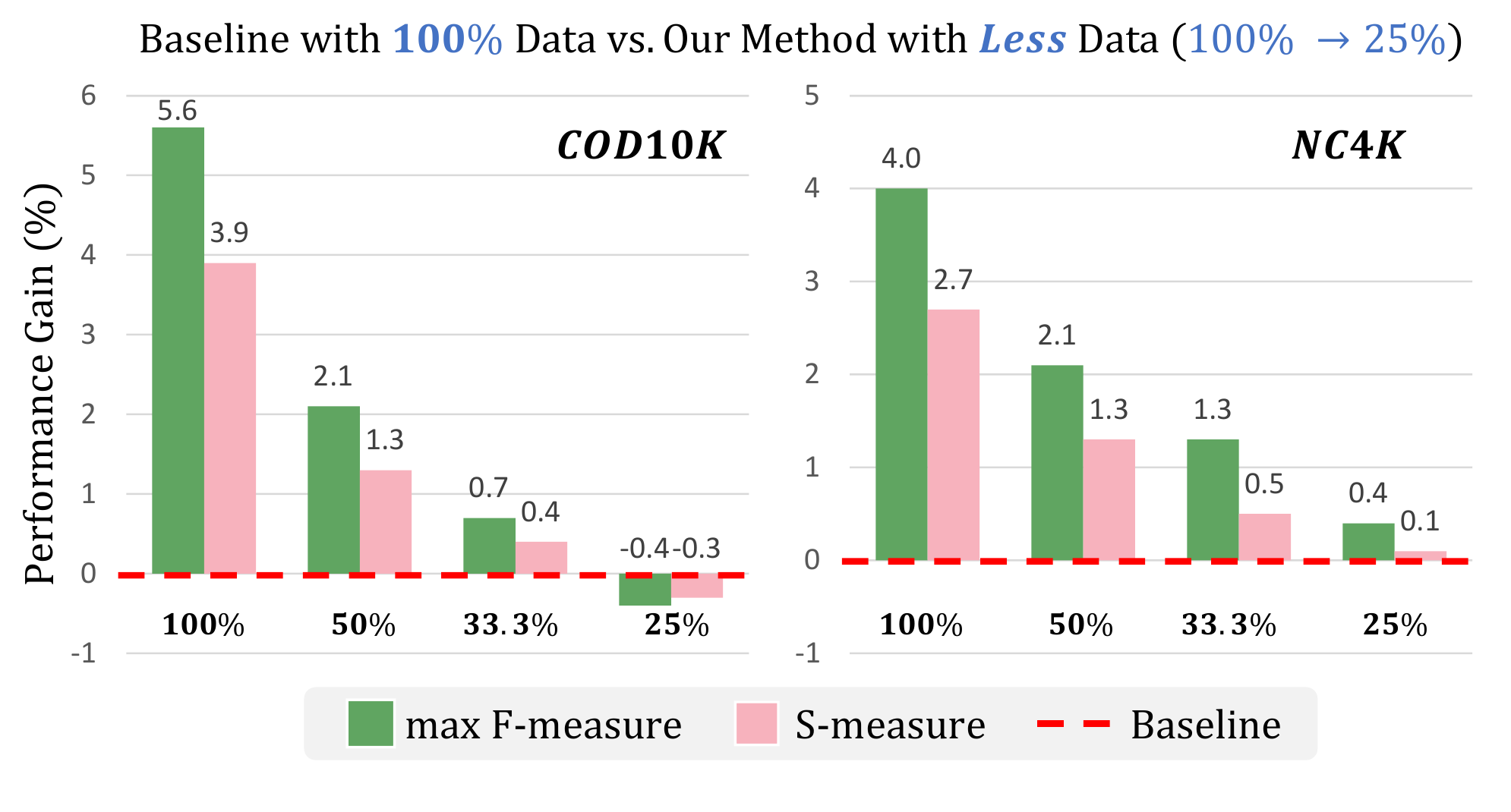}
\vspace{-8mm}
\caption{
Benefit of our method on \textbf{reducing training data}. When our network is trained only with 25\% data, the performance remains competitive compared to the baseline, \ie, the absolute performance gains are +0.4\% max F-measure and +0.1\% S-measure on NC4K compared to the baseline trained with all data.
}
\label{fig:training}
\vspace{-2mm}
\end{figure}

\noindent \textbf{Pop-out Under Reduced Training Data:} Here, we are interested in analyzing source-free depth's benefits. Therefore, we conduct different experiments by reducing the training data. As shown in Figure \ref{fig:training}(left), when both our PopNet and our baseline are trained with all data, our PopNet can lead to absolute improvements with 5.6\% in max F-measure and with 3.9\% in S-measure on COD10K dataset. While our PopNet is trained with only $25\%$ data, it can still achieve competitive performance compared to our baseline trained with all data. Similar phenomena can be observed in the NC4K dataset as shown in Figure \ref{fig:training}(right). To conclude, our method can efficiently explore the geometric prior and significantly reduce the required training data volume.

\begin{figure}[t]
\centering
\includegraphics[width=\linewidth,keepaspectratio]{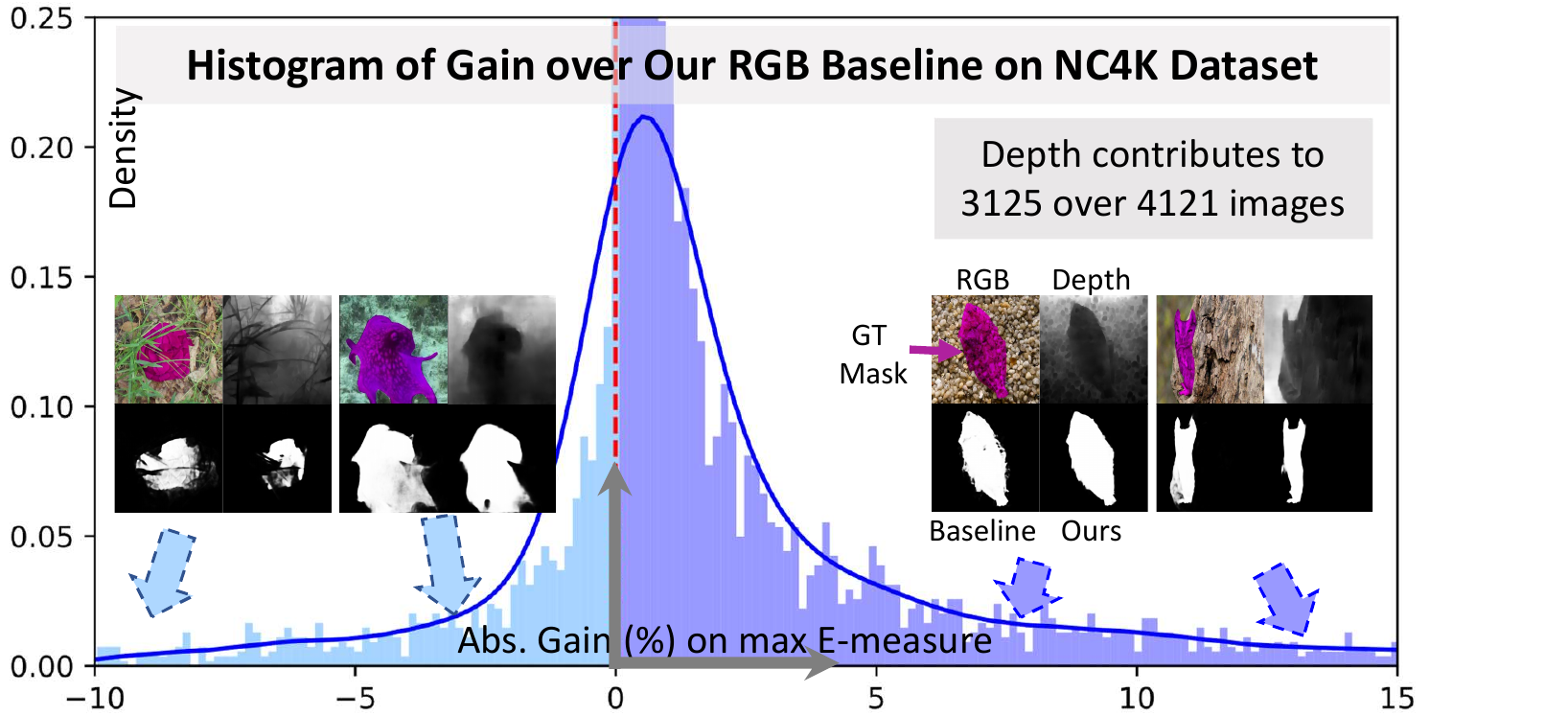}
\vspace{-6mm}
\caption{Histogram on the gain. Please zoom in for details.}
\label{fig:rebhist}
\vspace{-1mm}
\end{figure}

\noindent\textbf{Gain over baseline:} With depth cues, as shown in Figure \ref{fig:rebhist}, we boost the performance in 3125 over 4121 images ($\sim$75\% cases). We also show in Table \ref{tab:rebmultiobj} that our network performs favorably over baselines with single or multiple objects. Our method may fail when the source-free depth is clueless. This mainly happens when the object is well concealed and fools the depth network. However, such cases are also challenging, even for humans.  


\begin{table}[t]
\scriptsize	
\setlength\tabcolsep{0.6pt}
\renewcommand{\arraystretch}{1.0}
\begin{center}
\caption{Generalization and cost over different RGB-D baselines.}
\label{tab:general}
\begin{tabular}{  m{1.4cm}|| m{.75cm} m{.75cm} | m{.6cm} m{.6cm} m{.6cm} m{.6cm}| m{.6cm} m{.6cm} m{.6cm} m{.6cm} }
\hline

\hline

\hline
Dataset & Flops & Param &\multicolumn{4}{c|}{SIP \cite{fan2019rethinkingd3}} &  \multicolumn{4}{c}{NC4K \cite{lv2021simultaneously}} \\
\cline{4-11}  

Metric & (G) & (M) & 
        $M\downarrow$ & $F_{m}\uparrow$ &  $S_m\uparrow $ & $E_m\uparrow$ &
        $M\downarrow$ & $F_{m}\uparrow$ &  $S_m\uparrow $ & $E_m\uparrow$  \\
\hline

\hline

HAINet \cite{Li_2021_HAINet}  & 363.2 & 59.8 
                                      &  .053&   .899&   .874&   .919 
                                      &  .057&   .809&   .804&   .872 \\
\textbf{+ Ours} &373.7 & 72.5
                                      &  \textbf{.051}&   \textbf{.910}&   \textbf{.886}&   \textbf{.923}
                                      &  \textbf{.055}&   \textbf{.814}&   \textbf{.811}&   \textbf{.878} \\

\hline
SPNet \cite{zhouiccvspnet}  & 149.0 & 150.4
                                      & .044 & .911 & .887  &.914
                                      &  .054&   .828&   .825&   .874 \\
\textbf{+ Ours} & 159.5 & 163.1
                                      &  \textbf{.042}&   \textbf{.917}&   \textbf{.894}&   \textbf{.932}                         &  \textbf{.044}&   \textbf{.851}&   \textbf{.851}&   \textbf{.905} \\                    

\hline

\hline
\end{tabular}
\end{center}
\vspace{-5mm}
\end{table}

\begin{table}[t]
\scriptsize	
\setlength\tabcolsep{0.35pt}
\renewcommand{\arraystretch}{1.0}
\begin{center}
\caption{Multi-Object performance on NC4K \cite{lv2021simultaneously} with size $512^2$.}
\label{tab:rebmultiobj}
\begin{tabular}{  m{1.8cm} || m{.75cm}   m{.75cm} | m{.75cm} m{.75cm} | m{.75cm}   m{.75cm} | m{.75cm} m{.75cm}}
\hline

\hline

\hline

Obj. Nbr. (\%) & \multicolumn{2}{c|}{Single} (92\%)  & \multicolumn{2}{c|}{Two (6\%)} & \multicolumn{2}{c|}{More (2\%)} &
\multicolumn{2}{c}{Overall}\\

\cline{2-9}
 Metric &  $M\downarrow$ & $F_{m}\uparrow$ &  $M\downarrow$ & $F_{m}\uparrow$   &  $M\downarrow$ & $F_{m}\uparrow$   &  $M\downarrow$ & $F_{m}\uparrow$  
\\
\hline
RGB Baseline & .054 &.828 & .067 & .811 &.091 &.738  & .056 &.825 
\\
+ $D_{sf}$ & .050 & .842 &.063 &.821 &.093 &.746  & .051 &.839 
\\
\hdashline
\rowcolor[RGB]{235,235,250}  

\textbf{Ours} & \textbf{.040} & \textbf{.864} &\textbf{.051} &\textbf{.847} &\textbf{.079} &\textbf{.767}  & \textbf{.042} &\textbf{.861}
\\

\hline

\hline
\end{tabular}
\end{center}
\vspace{-6mm}
\end{table}

\label{dis}
\noindent\textbf{Are RGB-D Methods Better Than RGB-only Methods?} RGB-D methods are indeed better than RGB-only methods provided GT depth. However, only a few RGB-D methods can benefit from the source-free depth. For example, as shown in Table~\ref{tab:disc}, DASNet~\cite{zhao2020depth} achieves poorer performance compared to RGB baseline when trained with source-free depth. Similarly, even for one of the SOTA RGB-D models, SPNet~\cite{zhouiccvspnet}, the performance on NC4K dataset with RGB-only input is better than with additional source-free depth. Moreover, when provided with source-free depth, none of the existing RGB-D methods outperform the best performing RGB-only method (\eg, ZoomNet~\cite{pang2022zoom}) on COD. The same observation was also made on SOD as well. This could be because of the domain gap coupled with the fusion design, among others. We also found it non-trivial to extend the most well-performing RGB-only methods to the RGB-D case. Note that our PopNet performs better than all existing RGB-only and RGB-D methods, with source-free or GT depth maps.

\begin{table}[t]
\scriptsize	
\setlength\tabcolsep{0.5pt}
\renewcommand{\arraystretch}{1.0}
\begin{center}
\caption{Performance with RGB-$D_{sf}$ model vs. with RGB-only baseline. $D_{sf}$ stands for source-free depth.}
\label{tab:disc}
\begin{tabular}{  m{1.8cm}|| m{.76cm} m{.76cm} m{.76cm} m{.76cm}| m{.76cm} m{.76cm} m{.76cm} m{.76cm} }
\hline

\hline

\hline
Dataset &\multicolumn{4}{c|}{COD10K~\cite{fan2020camouflaged}} &  \multicolumn{4}{c}{NC4K~\cite{lv2021simultaneously}} \\

\cline{2-5} \cline{6-9}  

Metric & 
        $M\downarrow$ & $F_{m}\uparrow$ &  $S_m\uparrow $ & $E_m\uparrow$ &
        $M\downarrow$ & $F_{m}\uparrow$ &  $S_m\uparrow $ & $E_m\uparrow$  \\
\hline

DASNet~\cite{zhao2020depth}
&  \textbf{.041}&   \textbf{.643}&   .793&   \textbf{.864} &  \textbf{.055}&   \textbf{.747}&   \textbf{.830}&   \textbf{.879} \\
+ $D_{sf}$ &  \textbf{.041}&   .642&   \textbf{.796}&   .858 &  \textbf{.055}&   .743&   \textbf{.830}&   .874 \\
\hline

SPNet~\cite{zhouiccvspnet}
                                        &  .040&   .743&   .801&   .867 
                                      &  \textbf{.052}&   \textbf{.846}&   \textbf{.833}&   \textbf{.883} \\
+ $D_{sf}$
&  \textbf{.037}&   \textbf{.776}&  \textbf{.808}&   \textbf{.869} 
                                      &  .054&   .828&   .825&   .874 \\

\hline

\hline
\end{tabular}
\end{center}
\vspace{-5mm}

\end{table}

\section{Conclusion}
We demonstrate a successful case of cross-domain cross-task depth to semantics knowledge transfer using only the source model. In this paper, the source-free depth at the target offered by a given source model is used. The proposed method learns to transfer knowledge from depth to semantics using the objects' pop-out prior. We facilitate our network to use such prior by designing a novel network architecture. The designed network reasons about the objects by popping them out from the provided depth maps. This process is followed by separating objects from the background using the learned contact surface. We show that the joint learning of object pop-out and contact surface can be successfully supervised using the target semantics. Exhaustive experiments on SOD and COD benchmarks show the successful transfer of depth knowledge to the target, in terms of improved performance and generalization.

\noindent\textbf{Ackowledgement} The authors thank the anonymous reviewers and ACs for their tremendous efforts and helpful comments. This research is financed in part by the Conseil R\'egional de Bourgogne-Franche-Comt\'e, Toyota Motor Europe (research project TRACE-Zurich), the Alexander von Humboldt Foundation, and the Ministry of Education and Science of Bulgaria (support for INSAIT, part of the Bulgarian National Roadmap for Research Infrastructure).


{\small
\bibliographystyle{ieee_fullname}
\bibliography{egbib}
}

\end{document}